%% file: c_and_z.tex
\documentclass[acmtog]{acm}
\usepackage{color}
\usepackage{subfigure}
\usepackage{units}
\usepackage{expl3}
\usepackage{amsmath}
\usepackage{multirow}
\usepackage{colortbl}
\usepackage{algorithm}
\usepackage[noend]{algpseudocode}

\definecolor{Yellow}{rgb}{1,1, 0.6}
\definecolor{Red}{rgb}{1, 0.6, 0.6}
 
\usepackage{booktabs}
\setcitestyle{square}

\renewcommand\footnotetextcopyrightpermission[1]{}
\makeatletter
\renewcommand\@formatdoi[1]{\ignorespaces}
\makeatother

\begin{document}

\title{Segmentation of Objects by Hashing}

\author{J. D. Curt\'o$^{*,1,2,3,4}$, I. C. Zarza$^{*,1,2,3,4}$, A. Smola$^{3,5}$, and L. Gool$^{1}$.}
\affiliation{\institution{\\$^{1}$Eidgen\"ossische Technische Hochschule Z\"urich. $^{2}$The Chinese University of Hong Kong. \\ $^{3}$Carnegie Mellon. $^{4}$City University of Hong Kong. $^{5}$Amazon. \\ $^{*}$Both authors contributed equally.}}

\renewcommand{\shortauthors}{Curt\'o, Zarza, Smola and Gool.}
\renewcommand{\shorttitle}{Segmentation of Objects by Hashing.}

\authorsaddresses{\{curto,zarza,vangool\}@vision.ee.ethz.ch, smola@cs.cmu.edu \\ \href{https://www.decurto.tw}{decurto.tw} \href{https://www.dezarza.tw}{dezarza.tw}}

\begin{teaserfigure}
\centering
\includegraphics[width=\textwidth]{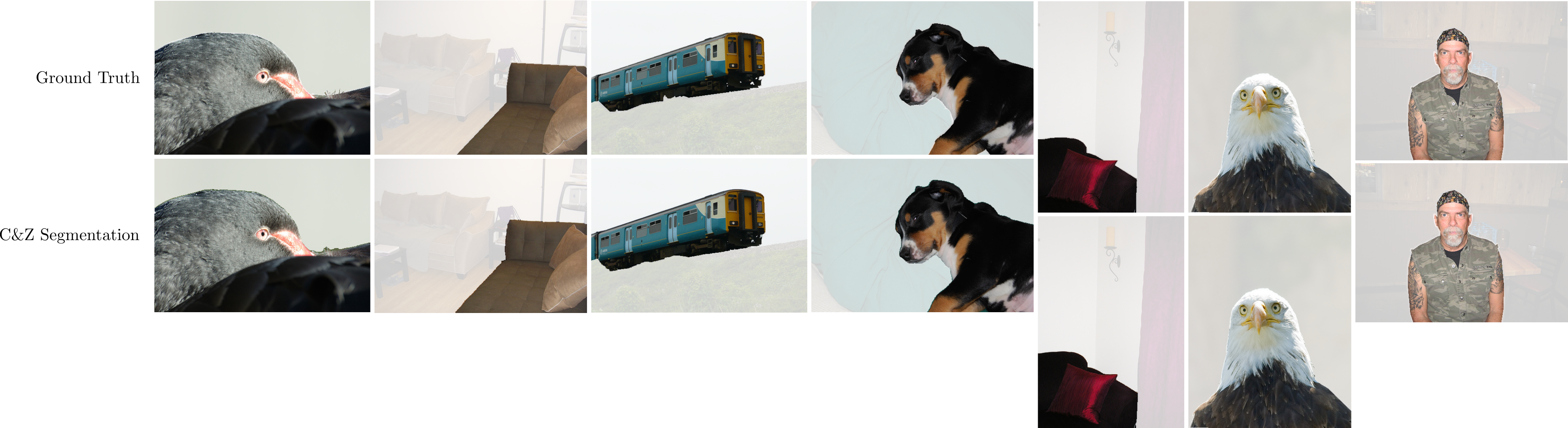}
   \caption{\textbf{Top Detections}. Top: VOC 2012 Ground Truth. Bottom: C\&Z Segmentation.}
\label{fgr:c_3}
\end{teaserfigure}

\ExplSyntaxOn
\newcommand\latinabbrev[1]{
  \peek_meaning:NTF . {
    #1\@}%
  { \peek_catcode:NTF a {
      #1.\@ }%
    {#1.\@}}}
\ExplSyntaxOff

\def\eg{e.g. }
\def\etal{et al. }


\begin{abstract}
We propose a novel approach to address the problem of Simultaneous Detection and Segmentation introduced in \cite{Hariharan14}. Using the hierarchical structures first presented in \cite{Arbelaez11} we use an efficient and accurate procedure that exploits the feature information of the hierarchy using Locality Sensitive Hashing. We build on recent work that utilizes convolutional neural networks to detect bounding boxes in an image \cite{Ren15} and then use the top similar hierarchical region that best fits each bounding box after hashing, we call this approach C\&Z Segmentation. We then refine our final segmentation results by automatic hierarchical pruning. C\&Z Segmentation introduces a train-free alternative to Hypercolumns \cite{Hariharan15}. We conduct extensive experiments on PASCAL VOC 2012 segmentation dataset, showing that C\&Z gives competitive state-of-the-art segmentations of objects.
\end{abstract}

\maketitle
\fancyfoot{}
\thispagestyle{empty}


\section{Introduction}
\label{sn:introduction}
Detection and Segmentation are key components in any toolbox of Computer Vision. In this paper we present a technique of hashing to segment an object given its bounding box and therefore attain simultaneously both Detection and Segmentation. At its heart lies a novel way to retrieve and generate a high-quality segmentation, which is crucial for a wide variety of CV applications. Simply put, we use a state-of-the-art convolutional network to detect the objects, but hashing on top of a high-quality hierarchy of regions to generate the segmentations. 
\\

Detection and Segmentation of Objects are two popular problems in Computer Vision and Machine Learning, historically treated as separated tasks. We consider these strongly related vision tasks as a unique one: detecting each object in an image and assigning to each pixel a binary label inside the corresponding bounding box.
\\

C\&Z Segmentation addresses the problem with a surprising different technique that deviates from the current norm of using proposal object candidates \cite{Arbelaez14}. In semantic segmentation the need for rich information models that entangle some kind of notion from the different parts that constitute an object is exacerbated.  To alleviate this issue we build on the use of the hierarchical model in \cite{Arbelaez11} and explore the rich space of information of the Ultrametric Contour Map (UCM) in order to find the best possible semantic segmentation of the given object. For this task, we exploit bounding boxes to facilitate the search. Hence, we simply hash the patches enclosed by the bounding boxes and retrieve closest nearest neighbors to the given objects, obtaining superior segmentations. Using this simple but effective technique we get the segmentation mask which is then refined using Hierarchical Section Pruning.
\\

We start from a detector of bounding boxes and refine the object support, as in Hypercolumns \cite{Hariharan15}. We propose here a train-free similarity hashing alternative to their approach.
 \\
 
We present a simple yet effective module that leverages the need for a training step and can provide segmentations after any given detector. Our approach is to use a state-of-the-art region-based CNN detector \cite{Ren15} as prior step to guide the process of segmentation. \\
 
\textbf{Outline:} We begin next with a high-level description of the proposed method and develop further the idea to propose Hierarchical Section Hashing and Hierarchical Section Pruning in Section \ref{sn:introduction} and Section \ref{sn:c}. Prior work follows in Section \ref{sn:work}. We conclude with the evaluation metrics in Section \ref{sn:e} and a brief discussion in Section \ref{sn:dn}.
\\

We start with a primer. C\&Z Segmentation consists on the following main blocks:

\begin{itemize}
\item \textbf{Detection of Objects using Bounding Boxes.}
We use a convolutional neural network \cite{Ren15} to detect all the objects in an image and generate the corresponding bounding boxes. We consider a detected object in an image as each output candidate thresholded by the class level score (benchmarks specifications in Section \ref{sn:e}).\\

\item \textbf{Hierarchy.}
The image is presented as a tree of hierarchical regions based on the UCM \cite{Arbelaez11}.\\

\item \textbf{Similarity Hashing.}
We develop Hierarchical Section Hashing based on the LSH technique of \cite{Charikar02}.\\

\item \textbf{Refinement of Regions.}
The segmentations are refined by the use of Hierarchical Section Pruning.\\

\item \textbf{Evaluation.}
We evaluate the results on the PASCAL VOC 2012 Segmentation dataset \cite{Everingham10} using the JACCARD index metric, which measures the average best overlap achieved by a segmentation mask for a ground truth object.
\end{itemize}

This work is inspired on how humans segment images: they first localize the objects they want to segment, they carefully inspect the object on the image by the use of their visual system and finally they choose the region that belongs to the body of that particular object. We believe that although the problem of detection has to be solved by the use of deep learning based on convolutional neural networks, in the same way that current breakthroughs have been attained in Generative Adversarial Networks (GAN) \cite{Goodfellow14,Radford16,Salimans16,Karras18,Curto17_2}, the problem of segmenting those objects is of a different nature and can be best understood by the use of hashing. Furthermore, current research trends link concepts of Deep Learning to Kernel Methods, proposing a unifying theory of learning in \cite{Curto17}.
\\

Our main contributions are presented as follows:
\begin{itemize}
\item Novel approach to solve the task of segmentation by similarity hashing exploiting the detection of objects using bounding boxes.\\

\item Use of hierarchical structures which are rich on semantic meaning instead of other current state-of-the-art techniques such as generation of proposal object candidates.\\

\item No need of training data for the task of segmentation under the framework of detection using bounding boxes, that is train-free accurate segmentations.\\

\item State-of-the-art results.
\end{itemize}

To our knowledge, we are the first to provide a segmentation based on hashing. This approach leverages the need to optimize over a high-dimensional space.
\\

Despite the success of region proposal methods in detection, they have in turn arisen as the main computational bottleneck of these approaches. Yet unlike the latter, hierarchical structures derived from the UCM are in comparison inexpensive to compute and store. While we continue to use a very fast region-based convolutional neural network (R-CNN) to solve the task of detection, we propose to solve the problem of segmentation by exploring efficiently the space generated by a hierarchical image.


\section{Prior Work}
\label{sn:work}
Recent works \cite{Hariharan14,He17} present Object Detection and Segmentation as a single problem. The task requires to detect and segment every instance of a category in the image. Our work is however more closely related to the approach in Hypercolumns \cite{Hariharan15}, where they go from bounding boxes to segmented masks. Our course of action is related in the sense that we propose an alternative that does not require a training step and can be used as an off-the-shelf high-quality segmenter. 
\\

For semantic segmentation \cite{Arbelaez11,Arbelaez14,Yu16, Chen17,Chen17_2, Chen18, Chen18_3}, there has been several approaches where they guide the process of segmentation by the use of a prior detector \cite{Girshick15,Ren15,Redmon16,Liu16,Redmon17,Huang17}. Recently, this strategy has also presented state-of-the-art results in person detection and pose estimation \cite{Papandreou17}. Alternate procedures count on a human-on-the-loop \cite{Acuna18}. Ongoing research on the matter uses Neural Architecture Search \cite{Zoph17,Zoph18,Liu18,Pham18,Zhang19,Li19} to design efficient architectures of neural networks for dense image prediction \cite{Chen18_2}. With the advent of present-day autonomous vehicles, the need to generate segmentations directly from the point cloud given by LIDAR \cite{Qi17,Qi17_2,Qi18}, as well as detect 3D objects \cite{Chen17_3,Zhou18,Yang18,Luo18,Liang18,Ku18,Lang18}, is also recently attracting lots of research efforts. Our segmenter starts rather than from raw pixels, \cite{Long15} and \cite{Badrinarayanan16}, or bounding box proposals as in \cite{Girshick14} and \cite{Hariharan14}, from a set of hierarchical regions given by the UCM structure. Other techniques rely on superpixels \eg \cite{Mostajabi15}. This is a distinct tactic that works directly on a different representation.


\section{C\&Z Segmentation}
\label{sn:c}
\begin{figure*}
\begin{center}
\includegraphics[scale=0.4]{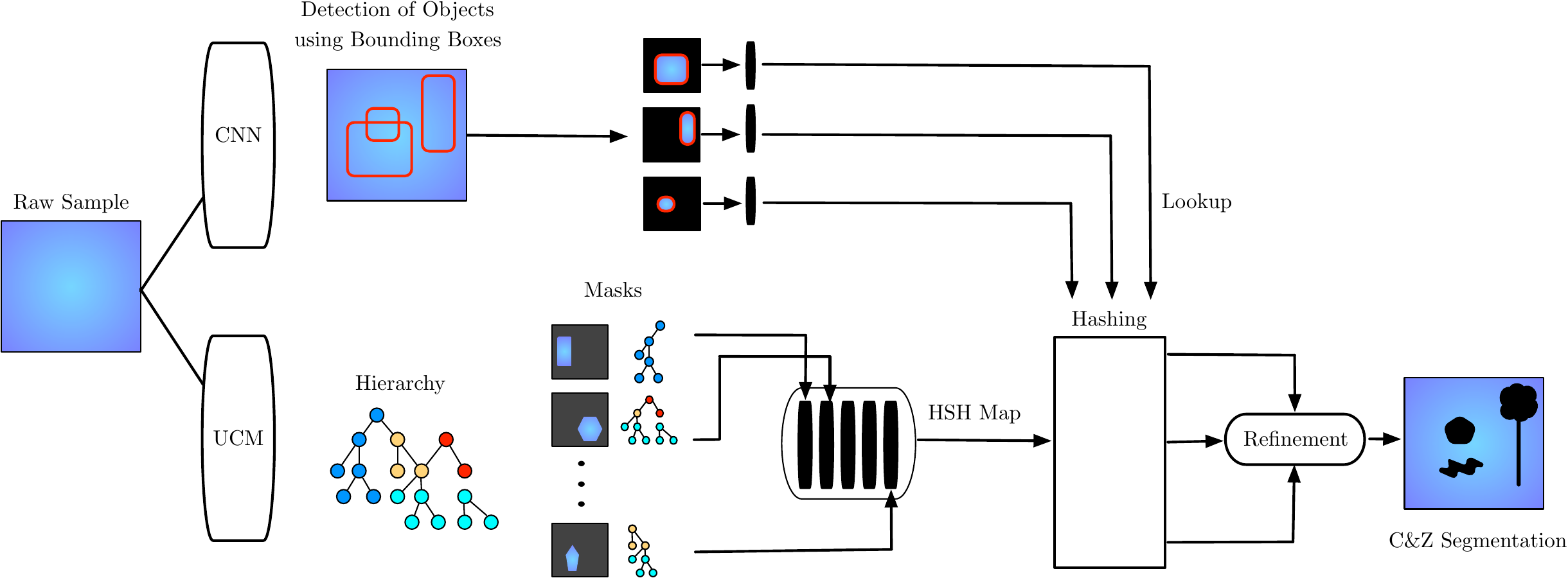}
\end{center}
   \caption{\textbf{C\&Z Segmentation}. We construct a hierarchical image based on the UCM and 'train' the HSH map by hashing each region of the parent partition nodes. To retrieve a segmentation mask, we 'test' the HSH map by doing a lookup of the detected area enclosed by the bounding box, videlicet a fast search of approximate nearest neighbors on the hierarchical structure, and finally refine the result through HSP.}
\label{fgr:c_2}
\end{figure*}

We delve into the details of the C\&Z Segmentation construction, Figure \ref{fgr:c_2}.

\subsection{Detection of Objects using Bounding Boxes}

\begin{figure}[H]
\begin{center}
\includegraphics[scale=0.24]{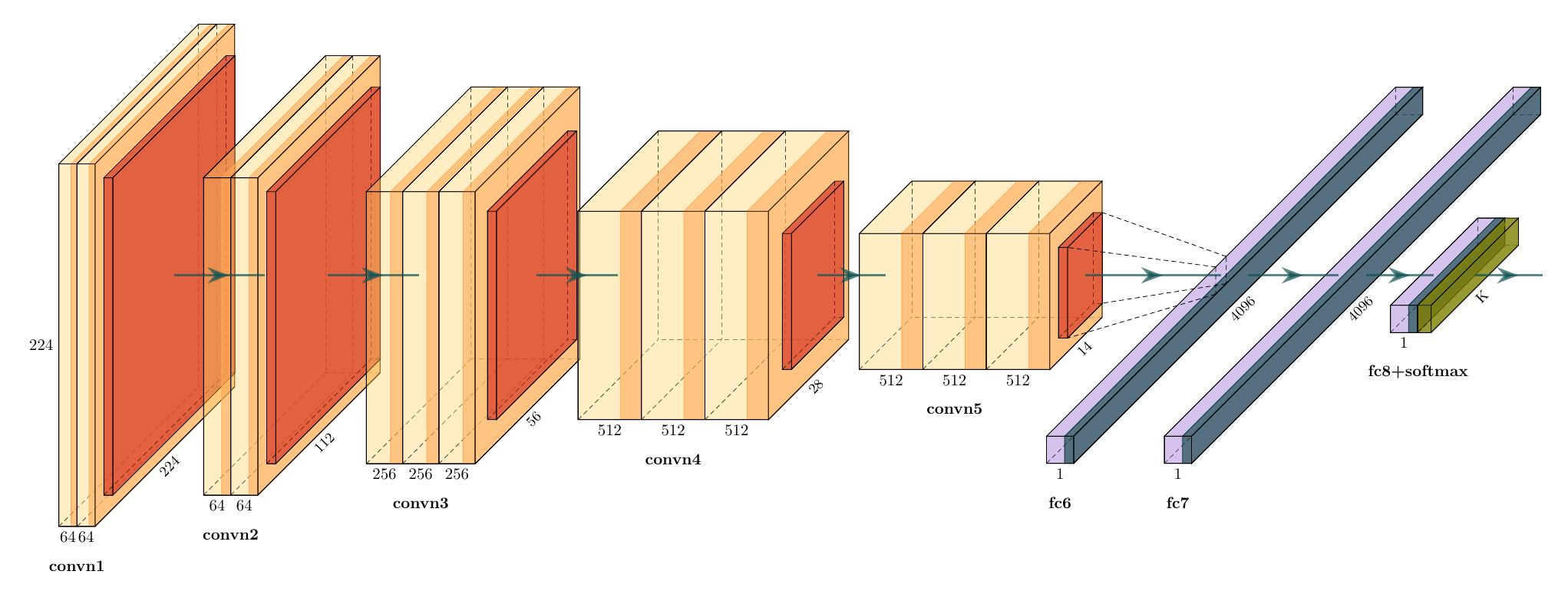}
\end{center}
   \caption{\textbf{VGG-16 Architecture}. VGG-16 model consists on an arrangement of convolutions, layers fully connected and softmax.}
\label{fgr:c_4}
\end{figure}

We begin by using the R-CNN object detector proposed by \cite{Ren15}, which is in turn based on \cite{Girshick15}. It introduces a Region Proposal Network (RPN) for the task of generating detection proposals and then solves the task of detection by the use of a FAST R-CNN detector. They train a CNN on ImageNet Classification and fine-tune the network on the VOC detection set. For our experiments, we use the network trained on VOC 2007 and 2012, and evaluate the results on the VOC 2012 evaluation set. We use the very deep VGG-16 model \cite{Simonyan15}, Figure \ref{fgr:c_4}.

\subsection{Hierarchical Image}

We consider the representation of a hierarchical image described in \cite{Arbelaez14}. Considering a segmentation of an image into regions that partition its domain $S=\{S_{c}\}_{c}$. A segmentation hierarchy is a family of partitions $\{S^{*},S^{1},\ldots,S^{L}\}$ such that: (1) $S^{*}$ is the finest set of superpixels, (2) $S^{L}$ is the complete domain, and (3) regions from coarse levels are unions of regions from fine levels.

\subsection{Hierarchical Section Hashing}

In this paper we introduce a novel segmentation algorithm that exploits bounding boxes to automatically select the best hierarchical region that segments the image. We introduce Hierarchical Section Hashing (HSH) which is in turn based on Locality-Sensitive Hashing (LSH). This algorithm helps us surpass the problem of computational complexity of the $k$-nearest neighbor rule and allows us to do a fast approximate neighbor search in the hierarchical structure of \cite{Arbelaez11}.
\\

HSH can be summarized as follows:
\begin{itemize}
\item Detect bounding boxes on an image using a state-of-the-art convolutional neural network \cite{Ren15}.\\
\item Construct a hierarchical image by using the UCM and convey the result as a hierarchical region tree.\\
\item Each hierarchical region is indexed by a number of tables of hashing using LSH and then constructing a HSH map. \\
\item Each bounding box is hashed into the HSH map to retrieve the approximate nearest neighbor in sublinear time.
\end{itemize}

C\&Z Segmentation has two main steps: first 'train' the HSH map with all the hierarchical regions of the image. Then 'test' the HSH map with all the detected bounding boxes to retrieve the approximate nearest neighbors that segment each of the objects in the image. The novelty of this approach is that it provides the best hierarchical region provided by the UCM structure that segments the object image. C\&Z exploits the detection of objects using bounding boxes because it relies on the correct detection of the object detector.

\subsection{Hierarchical Section Pruning}
The final piece is to refine the segmentations given by the HSH map by using what we call Hierarchical Section Pruning (HSP).
\\

HSP procedure can be summarized as follows:
\begin{itemize}
\item Once a segmentation mask has been selected for all the objects in the given image, and their bounding boxes recomputed, the bounding box overlap ratio for all box pair combinations, according to the intersection over union criterium, is performed.\\
\item Those masks that present overlap with other object masks on the same image are hierarchically unselected. We always proceed to unselect the low-level hierarchical regions, which by construction enclose a smaller region area and thus a single segmented object, from the high-level hierarchical region, which encloses more than one object and a bigger image area.\\
\item Finally, isolated pixels on the mask are erased to preserve a single connected segmentation.
\end{itemize}

HSP is based on the fact that each segmentation mask represents a node on the hierarchical region tree constructed from the UCM. Therefore, hierarchical sections containing more than one object represent higher level nodes on the hierarchy. When HSP is applied, low-level hierarchical regions are unselected from the high-level hierarchical sections and therefore replaced by mid-low level sections on the same region tree structure that represents a single object or a smaller area of the image. 
\\

\begin{figure}[H]
\begin{center}
   \includegraphics[scale=0.29]{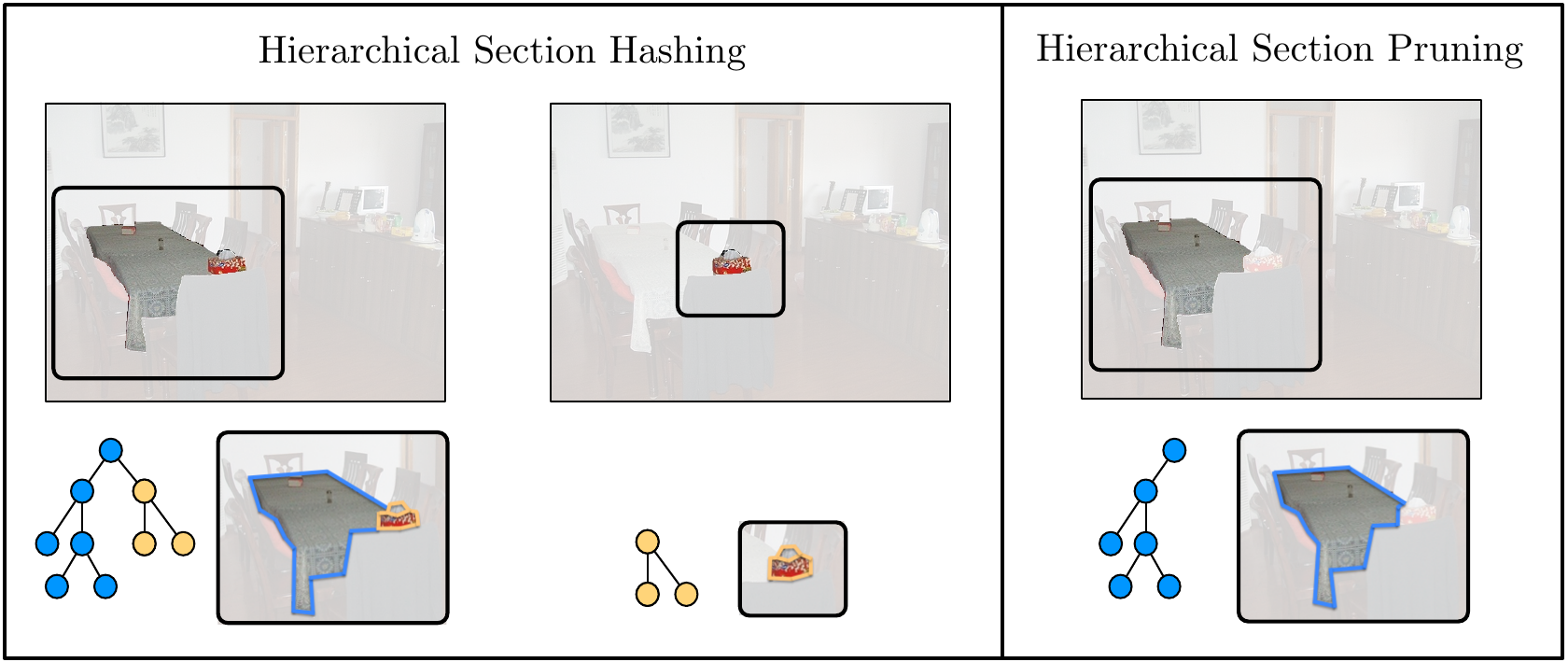}
\end{center}
   \caption{Left: HSH Visual Example. Right: HSP Visual Example.}
\label{fgr:c_1}
\end{figure}

HSH and HSP Visual Examples can be seen in Figure \ref{fgr:c_1}.

\begin{table*}
\begin{center}
\resizebox{\textwidth}{!}{%
\begin{tabular}{c c c c c c c c c c c c c c c c c c c c c c}
\hline
 & Aeroplane & Bicycle & Bird & Boat & Bottle & Bus & Car & Cat & Chair & Cow & Table & Dog & Horse & MBike & Person & Plant & Sheep & Sofa & Train & TV & Global\\
\hline\hline
\textbf{C\&Z Segmentation (Instance Level)} & 45.4 & 27.5 & 55.9 & 44.2 & 42.0 & 43.2 & 41.3 & 66.3 & 31.4 & 57.2 & 42.3 & 63.3 & 43.8 & 43.6 & 40.9 & 40.6 & 57.2 & 51.2 & 48.0 & 54.1 & 45.2\\
\hline\hline
\textbf{C\&Z Segmentation (Class Level)} & 33.3 & 18.5 & 48.1 & 37.5 & 40.7 & 45.1 & 39.4 & 59.9 & 23.3 & 51.0 & 43.3 & 60.4 & 39.8 & 43.1 & 34.6 & 37.2 & 51.0 & 47.0 & 53.6 & 54.2 & 43.1\\
\hline
\end{tabular}
}
\end{center}
\caption{\textbf{VOC 2012 Validation Set}. Per-class and global JACCARD Index Metric at instance level.}
\label{tbl:results}
\end{table*}

C\&Z Segmentation relies on the prior detection and therefore availability of bounding boxes 
for all the objects in a given image. The latter can be very useful as C\&Z can be understood as a simple and effective technique to provide high-quality segmentations of still images after any available detector of bounding boxes. Likewise, you get train-free off-the-shelf accurate segmentations for any given method that detects bounding boxes.

\subsection{Locality Sensitive Hashing}

Our goal is to retrieve the $k$-nearest neighbors of a given hierarchical vector, which we call \emph{image code}. In this setup we are limited by the curse of dimensionality and therefore using an exact search is inefficient. Our approach uses a technique of approximate nearest neighbors: Locality Sensitive Hashing (LSH). 
\\

A LSH function maps $x \rightarrow h(x)$ such that the similarity between $(\mathbf{x},\mathbf{y})$ is preserved as
\begin{align}
\bigg| \frac{d(h(\mathbf{x}),h(\mathbf{y}))}{D(\mathbf{x},\mathbf{y})} -1 \bigg| \leq \epsilon
\end{align}
which is not possible for all $D(\mathbf{x},\mathbf{y})$ but available for instance for euclidean metrics.
\\

We build on the LSH work of \cite{Indyk98,Gionis99,Charikar02,Shakhnarovich03,Shakhnarovich05,Andoni15}. LSH is a randomized hashing scheme, investigated with the primary goal of $\epsilon-R$ neighbor search. Its main constitutional block is a family of locality sensitive functions. We can define that a family $\mathcal{H}$ of functions $h:\mathcal{X} \rightarrow \{0,1\}$ is $(p_{1},p_{2},r,R)$-sensitive if, for any $\mathbf{x},\mathbf{y} \in \mathcal{X} $,
\begin{align}
Pr_{h\sim U[\mathcal{H}]} (h(\mathbf{x}) = h(\mathbf{y}) \ | \ || \mathbf{x} - \mathbf{y} || \leq r) \geq p_{1},\\
Pr_{h\sim U[\mathcal{H}]} (h(\mathbf{x}) = h(\mathbf{y}) \ | \ || \mathbf{x} - \mathbf{y} || \geq R) \leq p_{2},
\end{align}

\noindent
where these probabilities are chosen from a random choice of $h \in \mathcal{H}$. 
\\

Algorithm \ref{a:lsh} gives a simple description of the LSH algorithm for the given case when the distance of interest is $L_{1}$, which is the one in use in C\&Z Segmentation. The family $\mathcal{H}$ in this case contains axis-parallel stumps, which means the value of $h \in \mathcal{H}$ is generated by taking a simple dimension $d \in \{ 1, \ldots, \textnormal{dim}{(\mathcal{X}}) \}$ and thresholding it with some $T$:
\begin{align}
h^{LSH} = \left\{ \begin{array}{ll}
 1 & \textrm{if $x_{d} \leq T$},\\
 0 & \textrm{otherwise.}
  \end{array} \right.
\end{align}
 
 A LSH function $g: \mathcal{X} \rightarrow \{0, 1\}^{k}$ is formed by independently $k$ functions $h_{1},\ldots, h_{k} \in \mathcal{H}$. \\
 
That is, we can understand that an example in our hierarchical partition $S_{c} \in \mathcal{S}$ provides a $k$-bit key
 \begin{align}
 g(S_{c}) = [h_{1}(S_{c}), \ldots, h_{k}(S_{c})].
 \end{align}

This process is repeated $l$ times and produces $l$ independently constructed functions of hashing $g_{1}, \ldots, g_{l}$. The available reference ('training') data $S$ are indexed by each one of the $l$ functions of hashing, producing $l$ tables of hashing, namely each of the $S_{c}$ hierarchical partitions generated by all the corresponding parents of the hierarchical tree.

\begin{algorithm}[H]
\caption{LSH Algorithm \cite{Gionis99}}
\raggedright
 \textbf{Given:} Dataset $X = [\mathbf{x}_{1}, \mathbf{x}_{N}], \mathbf{x}_{c} \in \mathbb{R}^{\textnormal{dim}\mathcal(X)}.$\\
 \textbf{Given:} Number of bits $k$, number of tables $l$.\\
 \textbf{Output:} A set of
\begin{algorithmic}[1]
\For{$z = 1,\ldots, l$}
\For{$c = 1, \ldots, k$}
\State Randomly (uniformly) draw
\[d \in \{1, \ldots, \textnormal{dim}(\mathcal{X})\}.\]
\State Randomly (uniformly) draw
\[\min\{\mathbf{x}_{(d)}\} \leq v \leq \max\{\mathbf{x}_{(d)}\}.\]
\State Let $h_{c}^{z}$ be the function $\mathcal{X} \rightarrow \{0,1\}$ defined by
\begin{align*}
h_{c}^{z}(\mathbf{x}) = \left\{ \begin{array}{ll}
 1 & \textrm{if $x_{(d)} \leq v$},\\
 0 & \textrm{otherwise.}
  \end{array} \right.
\end{align*}
\EndFor
\State The $z$-th LSH function is $g_{z} = [h_{1}^{z}, \ldots, h_{k}^{z}]$.
\EndFor
\end{algorithmic}
\label{a:lsh}
\end{algorithm}
 
Once the LSH data structure has been built it can be used to perform a very efficient search for approximate neighbors in the following way.
When a query $S_{0}$ arrives, we compute its key for each table of hashing $z$, and record the examples $C=\{S_{1}^{l}, \ldots, S_{n_{l}}^{l}\}$ resulting from the lookup with that key. In other words, we find the 'training' examples that fell in the same bucket of the $l$-th table of hashing to which $S_{0}$ would fall. These $l$ lookup operations produce a set of candidate matches, $C=\cup_{z=1}^{l}C_{z}$. If this set is empty, the algorithm reports it and stops. Otherwise, the distances between candidate matches and $S_{0}$ are explicitly evaluated, and the examples that match the search criteria, which means that are closer to $S_{0}$ than $(1+\epsilon)R$, are returned.

\section{Evaluation and Results}
\label{sn:e}

We extensively evaluate C\&Z Segmentation on VOC 2012 validation set. Top detections from our algorithm can be seen in Figure \ref{fgr:c_3}.

\subsection{JACCARD Index Metric}

In Table \ref{tbl:results} we show the results of the JACCARD Index Metric. This measure represents the average best overlap achieved by a candidate for a ground truth object.
\\

C\&Z Segmentation with Jaccard at instance level $45.24$\% and Jaccard at class level $43.05$\%. Recall at overlap $0.5$ is $43.36$\%.


\section{Discussion}
\label{sn:dn}
In this paper we introduce C\&Z Segmentation, an algorithm to segment objects based on hashing that exploits the detection using bounding boxes. We show C\&Z achieves compelling results and generates off-the-shelf accurate segmentations.


\bibliographystyle{ACM-Reference-Format}
\input{c_and_z.bbl}

\end{document}

%% file: c_and_z.bbl

%% file: c_and_z.bbl
\begin{thebibliography}{53}


\ifx \showCODEN    \undefined \def \showCODEN     #1{\unskip}     \fi
\ifx \showDOI      \undefined \def \showDOI       #1{#1}\fi
\ifx \showISBNx    \undefined \def \showISBNx     #1{\unskip}     \fi
\ifx \showISBNxiii \undefined \def \showISBNxiii  #1{\unskip}     \fi
\ifx \showISSN     \undefined \def \showISSN      #1{\unskip}     \fi
\ifx \showLCCN     \undefined \def \showLCCN      #1{\unskip}     \fi
\ifx \shownote     \undefined \def \shownote      #1{#1}          \fi
\ifx \showarticletitle \undefined \def \showarticletitle #1{#1}   \fi
\ifx \showURL      \undefined \def \showURL       {\relax}        \fi
\providecommand\bibfield[2]{#2}
\providecommand\bibinfo[2]{#2}
\providecommand\natexlab[1]{#1}
\providecommand\showeprint[2][]{arXiv:#2}

\bibitem[\protect\citeauthoryear{Acuna, Ling, Kar, and Fidler}{Acuna
  et~al\mbox{.}}{2018}]%
        {Acuna18}
\bibfield{author}{\bibinfo{person}{D. Acuna}, \bibinfo{person}{H. Ling},
  \bibinfo{person}{A. Kar}, {and} \bibinfo{person}{S. Fidler}.}
  \bibinfo{year}{2018}\natexlab{}.
\newblock \showarticletitle{Efficient Interactive Annotation of Segmentation
  Datasets with Polygon-RNN++}.
\newblock \bibinfo{journal}{\emph{CVPR}} (\bibinfo{year}{2018}).
\newblock


\bibitem[\protect\citeauthoryear{Andoni, Indyk, Laarhoven, Razenshteyn, and
  Schmidt}{Andoni et~al\mbox{.}}{2015}]%
        {Andoni15}
\bibfield{author}{\bibinfo{person}{A. Andoni}, \bibinfo{person}{P. Indyk},
  \bibinfo{person}{T. Laarhoven}, \bibinfo{person}{I. Razenshteyn}, {and}
  \bibinfo{person}{L. Schmidt}.} \bibinfo{year}{2015}\natexlab{}.
\newblock \showarticletitle{Practical and Optimal LSH for Angular Distance}.
\newblock \bibinfo{journal}{\emph{NIPS}} (\bibinfo{year}{2015}).
\newblock


\bibitem[\protect\citeauthoryear{Arbel\'aez, Maire, Fowlkes, and
  Malik}{Arbel\'aez et~al\mbox{.}}{2011}]%
        {Arbelaez11}
\bibfield{author}{\bibinfo{person}{P. Arbel\'aez}, \bibinfo{person}{M. Maire},
  \bibinfo{person}{C. Fowlkes}, {and} \bibinfo{person}{J. Malik}.}
  \bibinfo{year}{2011}\natexlab{}.
\newblock \showarticletitle{Contour Detection and Hierarchical Image
  Segmentation}.
\newblock \bibinfo{journal}{\emph{TPAMI}} (\bibinfo{year}{2011}).
\newblock


\bibitem[\protect\citeauthoryear{Arbel\'aez, Pont-Tuset, Barron, Marques, and
  Malik}{Arbel\'aez et~al\mbox{.}}{2014}]%
        {Arbelaez14}
\bibfield{author}{\bibinfo{person}{P. Arbel\'aez}, \bibinfo{person}{J.
  Pont-Tuset}, \bibinfo{person}{J.~T. Barron}, \bibinfo{person}{F. Marques},
  {and} \bibinfo{person}{J. Malik}.} \bibinfo{year}{2014}\natexlab{}.
\newblock \showarticletitle{Multiscale Combinatorial Grouping}.
\newblock \bibinfo{journal}{\emph{CVPR}} (\bibinfo{year}{2014}).
\newblock


\bibitem[\protect\citeauthoryear{Badrinarayanan, Kendall, and
  Cipolla}{Badrinarayanan et~al\mbox{.}}{2016}]%
        {Badrinarayanan16}
\bibfield{author}{\bibinfo{person}{V. Badrinarayanan}, \bibinfo{person}{A.
  Kendall}, {and} \bibinfo{person}{R. Cipolla}.}
  \bibinfo{year}{2016}\natexlab{}.
\newblock \showarticletitle{{SegNet}: A Deep Convolutional Encoder-Decoder
  Architecture for Image Segmentation}.
\newblock \bibinfo{journal}{\emph{TPAMI}} (\bibinfo{year}{2016}).
\newblock


\bibitem[\protect\citeauthoryear{Charikar}{Charikar}{2002}]%
        {Charikar02}
\bibfield{author}{\bibinfo{person}{M. Charikar}.}
  \bibinfo{year}{2002}\natexlab{}.
\newblock \showarticletitle{Similarity Estimation Techniques from Rounding
  Algorithms}.
\newblock \bibinfo{journal}{\emph{STOC}} (\bibinfo{year}{2002}).
\newblock


\bibitem[\protect\citeauthoryear{Chen, Collins, Zhu, Papandreou, Zoph, Schroff,
  Adam, and Shlens}{Chen et~al\mbox{.}}{2018a}]%
        {Chen18_2}
\bibfield{author}{\bibinfo{person}{L. Chen}, \bibinfo{person}{M.~D. Collins},
  \bibinfo{person}{Y. Zhu}, \bibinfo{person}{G. Papandreou},
  \bibinfo{person}{B. Zoph}, \bibinfo{person}{F. Schroff}, \bibinfo{person}{H.
  Adam}, {and} \bibinfo{person}{J. Shlens}.} \bibinfo{year}{2018}\natexlab{a}.
\newblock \showarticletitle{Searching for Efficient Multi-Scale Architectures
  for Dense Image Prediction}.
\newblock \bibinfo{journal}{\emph{NIPS}} (\bibinfo{year}{2018}).
\newblock


\bibitem[\protect\citeauthoryear{Chen, Hermans, Papandreou, Schroff, Wang, and
  Adam}{Chen et~al\mbox{.}}{2018b}]%
        {Chen18_3}
\bibfield{author}{\bibinfo{person}{L. Chen}, \bibinfo{person}{A. Hermans},
  \bibinfo{person}{G. Papandreou}, \bibinfo{person}{F. Schroff},
  \bibinfo{person}{P. Wang}, {and} \bibinfo{person}{H. Adam}.}
  \bibinfo{year}{2018}\natexlab{b}.
\newblock \showarticletitle{{MaskLab}: Instance segmentation by refining object
  detection with semantic and direction features}.
\newblock \bibinfo{journal}{\emph{CVPR}} (\bibinfo{year}{2018}).
\newblock


\bibitem[\protect\citeauthoryear{Chen, Papandreou, Kokkinos, Murphy, and
  Yuille}{Chen et~al\mbox{.}}{2017b}]%
        {Chen17}
\bibfield{author}{\bibinfo{person}{L. Chen}, \bibinfo{person}{G. Papandreou},
  \bibinfo{person}{I. Kokkinos}, \bibinfo{person}{K. Murphy}, {and}
  \bibinfo{person}{A.~L. Yuille}.} \bibinfo{year}{2017}\natexlab{b}.
\newblock \showarticletitle{{DeepLab}: Semantic Image Segmentation with Deep
  Convolutional Nets, Atrous Convolution, and Fully Connected CRFs}.
\newblock \bibinfo{journal}{\emph{TPAMI}} (\bibinfo{year}{2017}).
\newblock


\bibitem[\protect\citeauthoryear{Chen, Papandreou, Schroff, and Adam}{Chen
  et~al\mbox{.}}{2017c}]%
        {Chen17_2}
\bibfield{author}{\bibinfo{person}{L. Chen}, \bibinfo{person}{G. Papandreou},
  \bibinfo{person}{F. Schroff}, {and} \bibinfo{person}{H. Adam}.}
  \bibinfo{year}{2017}\natexlab{c}.
\newblock \showarticletitle{Rethinking Atrous Convolution for Semantic Image
  Segmentation}.
\newblock \bibinfo{journal}{\emph{arXiv:1706.05587}} (\bibinfo{year}{2017}).
\newblock


\bibitem[\protect\citeauthoryear{Chen, Zhu, Papandreou, Schroff, and Adam}{Chen
  et~al\mbox{.}}{2018c}]%
        {Chen18}
\bibfield{author}{\bibinfo{person}{L. Chen}, \bibinfo{person}{Y. Zhu},
  \bibinfo{person}{G. Papandreou}, \bibinfo{person}{F. Schroff}, {and}
  \bibinfo{person}{H. Adam}.} \bibinfo{year}{2018}\natexlab{c}.
\newblock \showarticletitle{Encoder-Decoder with Atrous Separable Convolution
  for Semantic Image Segmentation}.
\newblock \bibinfo{journal}{\emph{ECCV}} (\bibinfo{year}{2018}).
\newblock


\bibitem[\protect\citeauthoryear{Chen, Ma, Wan, Li, and Xia}{Chen
  et~al\mbox{.}}{2017a}]%
        {Chen17_3}
\bibfield{author}{\bibinfo{person}{X. Chen}, \bibinfo{person}{H. Ma},
  \bibinfo{person}{J. Wan}, \bibinfo{person}{B. Li}, {and} \bibinfo{person}{T.
  Xia}.} \bibinfo{year}{2017}\natexlab{a}.
\newblock \showarticletitle{Multi-View {3D} Object Detection Network for
  Autonomous Driving}.
\newblock \bibinfo{journal}{\emph{CVPR}} (\bibinfo{year}{2017}).
\newblock


\bibitem[\protect\citeauthoryear{Curt\'o, Zarza, Torre, King, and Lyu}{Curt\'o
  et~al\mbox{.}}{2017a}]%
        {Curto17_2}
\bibfield{author}{\bibinfo{person}{J.~D. Curt\'o}, \bibinfo{person}{I.~C.
  Zarza}, \bibinfo{person}{F. Torre}, \bibinfo{person}{I. King}, {and}
  \bibinfo{person}{M.~R. Lyu}.} \bibinfo{year}{2017}\natexlab{a}.
\newblock \showarticletitle{High-resolution Deep Convolutional Generative
  Adversarial Networks}.
\newblock \bibinfo{journal}{\emph{arXiv:1711.06491}} (\bibinfo{year}{2017}).
\newblock


\bibitem[\protect\citeauthoryear{Curt\'o, Zarza, Yang, Smola, Torre, Ngo, and
  Gool}{Curt\'o et~al\mbox{.}}{2017b}]%
        {Curto17}
\bibfield{author}{\bibinfo{person}{J.~D. Curt\'o}, \bibinfo{person}{I.~C.
  Zarza}, \bibinfo{person}{F. Yang}, \bibinfo{person}{A. Smola},
  \bibinfo{person}{F. Torre}, \bibinfo{person}{C.~W. Ngo}, {and}
  \bibinfo{person}{L. Gool}.} \bibinfo{year}{2017}\natexlab{b}.
\newblock \showarticletitle{{McKernel}: A Library for Approximate Kernel
  Expansions in Log-linear Time.}
\newblock \bibinfo{journal}{\emph{arXiv:1702.08159}} (\bibinfo{year}{2017}).
\newblock


\bibitem[\protect\citeauthoryear{Everingham, Gool, Williams, Winn, and
  Zisserman}{Everingham et~al\mbox{.}}{2010}]%
        {Everingham10}
\bibfield{author}{\bibinfo{person}{M. Everingham}, \bibinfo{person}{L. Gool},
  \bibinfo{person}{C.~K.~I. Williams}, \bibinfo{person}{J. Winn}, {and}
  \bibinfo{person}{A. Zisserman}.} \bibinfo{year}{2010}\natexlab{}.
\newblock \showarticletitle{The Pascal Visual Object Classes ({VOC})
  Challenge}.
\newblock \bibinfo{journal}{\emph{IJCV}} (\bibinfo{year}{2010}).
\newblock


\bibitem[\protect\citeauthoryear{Gionis, Indyk, and Motwani}{Gionis
  et~al\mbox{.}}{1999}]%
        {Gionis99}
\bibfield{author}{\bibinfo{person}{A. Gionis}, \bibinfo{person}{P. Indyk},
  {and} \bibinfo{person}{R. Motwani}.} \bibinfo{year}{1999}\natexlab{}.
\newblock \showarticletitle{Similarity Search in High Dimensions via Hashing}.
\newblock \bibinfo{journal}{\emph{VLDB}} (\bibinfo{year}{1999}).
\newblock


\bibitem[\protect\citeauthoryear{Girshick}{Girshick}{2015}]%
        {Girshick15}
\bibfield{author}{\bibinfo{person}{R. Girshick}.}
  \bibinfo{year}{2015}\natexlab{}.
\newblock \showarticletitle{FAST R-CNN}.
\newblock \bibinfo{journal}{\emph{ICCV}} (\bibinfo{year}{2015}).
\newblock


\bibitem[\protect\citeauthoryear{Girshick, Donahue, Darrell, and
  Malik}{Girshick et~al\mbox{.}}{2014}]%
        {Girshick14}
\bibfield{author}{\bibinfo{person}{R. Girshick}, \bibinfo{person}{J. Donahue},
  \bibinfo{person}{T. Darrell}, {and} \bibinfo{person}{J. Malik}.}
  \bibinfo{year}{2014}\natexlab{}.
\newblock \showarticletitle{Rich feature hierarchies for accurate object
  detection and semantic segmentation}.
\newblock \bibinfo{journal}{\emph{CVPR}} (\bibinfo{year}{2014}).
\newblock


\bibitem[\protect\citeauthoryear{Goodfellow, Pouget-Abadie, Mirza, Xu,
  Warde-Farley, Ozair, Courville, and Bengio}{Goodfellow et~al\mbox{.}}{2014}]%
        {Goodfellow14}
\bibfield{author}{\bibinfo{person}{I. Goodfellow}, \bibinfo{person}{J.
  Pouget-Abadie}, \bibinfo{person}{M. Mirza}, \bibinfo{person}{B. Xu},
  \bibinfo{person}{D. Warde-Farley}, \bibinfo{person}{S. Ozair},
  \bibinfo{person}{A. Courville}, {and} \bibinfo{person}{Y. Bengio}.}
  \bibinfo{year}{2014}\natexlab{}.
\newblock \showarticletitle{Generative Adversarial Neworks}.
\newblock \bibinfo{journal}{\emph{NIPS}} (\bibinfo{year}{2014}).
\newblock


\bibitem[\protect\citeauthoryear{Hariharan, Arbel\'aez, Girshick, and
  Malik}{Hariharan et~al\mbox{.}}{2014}]%
        {Hariharan14}
\bibfield{author}{\bibinfo{person}{B. Hariharan}, \bibinfo{person}{P.
  Arbel\'aez}, \bibinfo{person}{R. Girshick}, {and} \bibinfo{person}{J.
  Malik}.} \bibinfo{year}{2014}\natexlab{}.
\newblock \showarticletitle{Simultaneous Detection and Segmentation}.
\newblock \bibinfo{journal}{\emph{ECCV}} (\bibinfo{year}{2014}).
\newblock


\bibitem[\protect\citeauthoryear{Hariharan, Arbel\'aez, Girshick, and
  Malik}{Hariharan et~al\mbox{.}}{2015}]%
        {Hariharan15}
\bibfield{author}{\bibinfo{person}{B. Hariharan}, \bibinfo{person}{P.
  Arbel\'aez}, \bibinfo{person}{R. Girshick}, {and} \bibinfo{person}{J.
  Malik}.} \bibinfo{year}{2015}\natexlab{}.
\newblock \showarticletitle{Hypercolumns for Object Segmentation and
  Fine-grained Localization}.
\newblock \bibinfo{journal}{\emph{CVPR}} (\bibinfo{year}{2015}).
\newblock


\bibitem[\protect\citeauthoryear{He, Gkioxari, Doll\'ar, and Girshick}{He
  et~al\mbox{.}}{2017}]%
        {He17}
\bibfield{author}{\bibinfo{person}{K. He}, \bibinfo{person}{G. Gkioxari},
  \bibinfo{person}{P. Doll\'ar}, {and} \bibinfo{person}{R. Girshick}.}
  \bibinfo{year}{2017}\natexlab{}.
\newblock \showarticletitle{MASK R-CNN}.
\newblock \bibinfo{journal}{\emph{ICCV}} (\bibinfo{year}{2017}).
\newblock


\bibitem[\protect\citeauthoryear{Huang, Rathod, Sun, Zhu, Korattikara, Fathi,
  Fischer, Wojna, Song, Guadarrama, and Murphy}{Huang et~al\mbox{.}}{2017}]%
        {Huang17}
\bibfield{author}{\bibinfo{person}{J. Huang}, \bibinfo{person}{V. Rathod},
  \bibinfo{person}{C. Sun}, \bibinfo{person}{M. Zhu}, \bibinfo{person}{A.
  Korattikara}, \bibinfo{person}{A. Fathi}, \bibinfo{person}{I. Fischer},
  \bibinfo{person}{Z. Wojna}, \bibinfo{person}{Y. Song}, \bibinfo{person}{S.
  Guadarrama}, {and} \bibinfo{person}{K. Murphy}.}
  \bibinfo{year}{2017}\natexlab{}.
\newblock \showarticletitle{Speed/accuracy trade-offs for modern convolutional
  object detectors}.
\newblock \bibinfo{journal}{\emph{CVPR}} (\bibinfo{year}{2017}).
\newblock


\bibitem[\protect\citeauthoryear{Indyk and Motwani}{Indyk and Motwani}{1998}]%
        {Indyk98}
\bibfield{author}{\bibinfo{person}{P. Indyk} {and} \bibinfo{person}{R.
  Motwani}.} \bibinfo{year}{1998}\natexlab{}.
\newblock \showarticletitle{Approximate nearest neighbors: towards removing the
  curse of dimensionality}.
\newblock \bibinfo{journal}{\emph{STOC}} (\bibinfo{year}{1998}).
\newblock


\bibitem[\protect\citeauthoryear{Karras, Aila, Laine, and Lehtinen}{Karras
  et~al\mbox{.}}{2018}]%
        {Karras18}
\bibfield{author}{\bibinfo{person}{T. Karras}, \bibinfo{person}{T. Aila},
  \bibinfo{person}{S. Laine}, {and} \bibinfo{person}{J. Lehtinen}.}
  \bibinfo{year}{2018}\natexlab{}.
\newblock \showarticletitle{Progressive Growing of GANs for Improved Quality,
  Stability, and Variation}.
\newblock \bibinfo{journal}{\emph{ICLR}} (\bibinfo{year}{2018}).
\newblock


\bibitem[\protect\citeauthoryear{Ku, Mozifian, Lee, Harakeh, and Waslander}{Ku
  et~al\mbox{.}}{2018}]%
        {Ku18}
\bibfield{author}{\bibinfo{person}{J. Ku}, \bibinfo{person}{M. Mozifian},
  \bibinfo{person}{J. Lee}, \bibinfo{person}{A. Harakeh}, {and}
  \bibinfo{person}{S. Waslander}.} \bibinfo{year}{2018}\natexlab{}.
\newblock \showarticletitle{Joint {3D} Proposal Generation and Object Detection
  from View Aggregation}.
\newblock \bibinfo{journal}{\emph{IROS}} (\bibinfo{year}{2018}).
\newblock


\bibitem[\protect\citeauthoryear{Lang, Vora, Caesar, Zhou, Yang, and
  Beijbom}{Lang et~al\mbox{.}}{2018}]%
        {Lang18}
\bibfield{author}{\bibinfo{person}{A.~H. Lang}, \bibinfo{person}{S. Vora},
  \bibinfo{person}{H. Caesar}, \bibinfo{person}{L. Zhou}, \bibinfo{person}{J.
  Yang}, {and} \bibinfo{person}{O. Beijbom}.} \bibinfo{year}{2018}\natexlab{}.
\newblock \showarticletitle{{PointPillars}: Fast Encoders for Object Detection
  from Point Clouds}.
\newblock \bibinfo{journal}{\emph{arXiv:1812.05784}} (\bibinfo{year}{2018}).
\newblock


\bibitem[\protect\citeauthoryear{Li and Talwalkar}{Li and Talwalkar}{2019}]%
        {Li19}
\bibfield{author}{\bibinfo{person}{L. Li} {and} \bibinfo{person}{A.
  Talwalkar}.} \bibinfo{year}{2019}\natexlab{}.
\newblock \showarticletitle{Random Search and Reproducibility for Neural
  Architecture Search}.
\newblock \bibinfo{journal}{\emph{arXiv:1902.07638}} (\bibinfo{year}{2019}).
\newblock


\bibitem[\protect\citeauthoryear{Liang, Yang, Wang, and Urtasun}{Liang
  et~al\mbox{.}}{2018}]%
        {Liang18}
\bibfield{author}{\bibinfo{person}{M. Liang}, \bibinfo{person}{B. Yang},
  \bibinfo{person}{S. Wang}, {and} \bibinfo{person}{R. Urtasun}.}
  \bibinfo{year}{2018}\natexlab{}.
\newblock \showarticletitle{Deep Continuous Fusion for Multi-Sensor {3D} Object
  Detection}.
\newblock \bibinfo{journal}{\emph{ECCV}} (\bibinfo{year}{2018}).
\newblock


\bibitem[\protect\citeauthoryear{Liu, Zoph, Neumann, Shlens, Hua, Li, Fei-Fei,
  Yuille, Huang, and Murphy}{Liu et~al\mbox{.}}{2018}]%
        {Liu18}
\bibfield{author}{\bibinfo{person}{C. Liu}, \bibinfo{person}{B. Zoph},
  \bibinfo{person}{M. Neumann}, \bibinfo{person}{J. Shlens},
  \bibinfo{person}{W. Hua}, \bibinfo{person}{L.-J. Li}, \bibinfo{person}{L.
  Fei-Fei}, \bibinfo{person}{A. Yuille}, \bibinfo{person}{J. Huang}, {and}
  \bibinfo{person}{K. Murphy}.} \bibinfo{year}{2018}\natexlab{}.
\newblock \showarticletitle{Progressive neural architecture search}.
\newblock \bibinfo{journal}{\emph{ECCV}} (\bibinfo{year}{2018}).
\newblock


\bibitem[\protect\citeauthoryear{Liu, Anguelov, Erhan, Szegedy, Reed, Fu, and
  Berg}{Liu et~al\mbox{.}}{2016}]%
        {Liu16}
\bibfield{author}{\bibinfo{person}{W. Liu}, \bibinfo{person}{D. Anguelov},
  \bibinfo{person}{D. Erhan}, \bibinfo{person}{C. Szegedy}, \bibinfo{person}{S.
  Reed}, \bibinfo{person}{C. Fu}, {and} \bibinfo{person}{A.~C. Berg}.}
  \bibinfo{year}{2016}\natexlab{}.
\newblock \showarticletitle{{SSD}: Single Shot MultiBox Detector}.
\newblock \bibinfo{journal}{\emph{ECCV}} (\bibinfo{year}{2016}).
\newblock


\bibitem[\protect\citeauthoryear{Long, Shelhamer, and Darrell}{Long
  et~al\mbox{.}}{2015}]%
        {Long15}
\bibfield{author}{\bibinfo{person}{J. Long}, \bibinfo{person}{E. Shelhamer},
  {and} \bibinfo{person}{T. Darrell}.} \bibinfo{year}{2015}\natexlab{}.
\newblock \showarticletitle{Fully Convolutional Networks for Semantic
  Segmentation}.
\newblock \bibinfo{journal}{\emph{CVPR}} (\bibinfo{year}{2015}).
\newblock


\bibitem[\protect\citeauthoryear{Luo, Yang, and Urtasun}{Luo
  et~al\mbox{.}}{2018}]%
        {Luo18}
\bibfield{author}{\bibinfo{person}{W. Luo}, \bibinfo{person}{B. Yang}, {and}
  \bibinfo{person}{R. Urtasun}.} \bibinfo{year}{2018}\natexlab{}.
\newblock \showarticletitle{Fast and Furious: Real Time End-to-End {3D}
  Detection, Tracking and Motion Forecasting with a Single Convolutional Net}.
\newblock \bibinfo{journal}{\emph{CVPR}} (\bibinfo{year}{2018}).
\newblock


\bibitem[\protect\citeauthoryear{Mostajabi, Yadollahpour, and
  Shakhnarovich}{Mostajabi et~al\mbox{.}}{2015}]%
        {Mostajabi15}
\bibfield{author}{\bibinfo{person}{M. Mostajabi}, \bibinfo{person}{P.
  Yadollahpour}, {and} \bibinfo{person}{G. Shakhnarovich}.}
  \bibinfo{year}{2015}\natexlab{}.
\newblock \showarticletitle{Feedforward semantic segmentation with zoom-out
  features}.
\newblock \bibinfo{journal}{\emph{CVPR}} (\bibinfo{year}{2015}).
\newblock


\bibitem[\protect\citeauthoryear{Papandreou, Zhu, Kanazawa, Toshev, Tompson,
  Bregler, and Murphy}{Papandreou et~al\mbox{.}}{2017}]%
        {Papandreou17}
\bibfield{author}{\bibinfo{person}{G. Papandreou}, \bibinfo{person}{T. Zhu},
  \bibinfo{person}{N. Kanazawa}, \bibinfo{person}{A. Toshev},
  \bibinfo{person}{J. Tompson}, \bibinfo{person}{C. Bregler}, {and}
  \bibinfo{person}{K. Murphy}.} \bibinfo{year}{2017}\natexlab{}.
\newblock \showarticletitle{Towards Accurate Multi-person Pose Estimation in
  the Wild}.
\newblock \bibinfo{journal}{\emph{CVPR}} (\bibinfo{year}{2017}).
\newblock


\bibitem[\protect\citeauthoryear{Pham, Guan, Zoph, Le, and Dean}{Pham
  et~al\mbox{.}}{2018}]%
        {Pham18}
\bibfield{author}{\bibinfo{person}{H. Pham}, \bibinfo{person}{M.~Y. Guan},
  \bibinfo{person}{B. Zoph}, \bibinfo{person}{Q.~V. Le}, {and}
  \bibinfo{person}{J. Dean}.} \bibinfo{year}{2018}\natexlab{}.
\newblock \showarticletitle{Efficient Neural Architecture Search via Parameter
  Sharing}.
\newblock \bibinfo{journal}{\emph{ICML}} (\bibinfo{year}{2018}).
\newblock


\bibitem[\protect\citeauthoryear{Qi, Liu, Wu, Su, and Guibas}{Qi
  et~al\mbox{.}}{2018}]%
        {Qi18}
\bibfield{author}{\bibinfo{person}{C.~R. Qi}, \bibinfo{person}{W. Liu},
  \bibinfo{person}{C. Wu}, \bibinfo{person}{H. Su}, {and}
  \bibinfo{person}{L.~J. Guibas}.} \bibinfo{year}{2018}\natexlab{}.
\newblock \showarticletitle{Frustum {PointNets} for {3D} Object Detection from
  RGB-D Data}.
\newblock \bibinfo{journal}{\emph{CVPR}} (\bibinfo{year}{2018}).
\newblock


\bibitem[\protect\citeauthoryear{Qi, Su, Mo, and Guibas}{Qi
  et~al\mbox{.}}{2017a}]%
        {Qi17}
\bibfield{author}{\bibinfo{person}{C.~R. Qi}, \bibinfo{person}{H. Su},
  \bibinfo{person}{K. Mo}, {and} \bibinfo{person}{L.~J. Guibas}.}
  \bibinfo{year}{2017}\natexlab{a}.
\newblock \showarticletitle{{PointNet}: Deep Learning on Point Sets for {3D}
  Classification and Segmentation}.
\newblock \bibinfo{journal}{\emph{CVPR}} (\bibinfo{year}{2017}).
\newblock


\bibitem[\protect\citeauthoryear{Qi, Yi, Su, and Guibas}{Qi
  et~al\mbox{.}}{2017b}]%
        {Qi17_2}
\bibfield{author}{\bibinfo{person}{C.~R. Qi}, \bibinfo{person}{L. Yi},
  \bibinfo{person}{H. Su}, {and} \bibinfo{person}{L.~J. Guibas}.}
  \bibinfo{year}{2017}\natexlab{b}.
\newblock \showarticletitle{{PointNet++}: Deep Hierarchical Feature Learning on
  Point Sets in a Metric Space}.
\newblock \bibinfo{journal}{\emph{NIPS}} (\bibinfo{year}{2017}).
\newblock


\bibitem[\protect\citeauthoryear{Radford, Metz, and Chintala}{Radford
  et~al\mbox{.}}{2016}]%
        {Radford16}
\bibfield{author}{\bibinfo{person}{A. Radford}, \bibinfo{person}{L. Metz},
  {and} \bibinfo{person}{S. Chintala}.} \bibinfo{year}{2016}\natexlab{}.
\newblock \showarticletitle{Unsupervised Representation Learning with Deep
  Convolutional Generative Adversarial Network}.
\newblock \bibinfo{journal}{\emph{ICLR}} (\bibinfo{year}{2016}).
\newblock


\bibitem[\protect\citeauthoryear{Redmon, Divvala, Girshick, and Farhadi}{Redmon
  et~al\mbox{.}}{2016}]%
        {Redmon16}
\bibfield{author}{\bibinfo{person}{J. Redmon}, \bibinfo{person}{S. Divvala},
  \bibinfo{person}{R. Girshick}, {and} \bibinfo{person}{A. Farhadi}.}
  \bibinfo{year}{2016}\natexlab{}.
\newblock \showarticletitle{You Only Look Once: Unified, Real-Time Object
  Detection}.
\newblock \bibinfo{journal}{\emph{CVPR}} (\bibinfo{year}{2016}).
\newblock


\bibitem[\protect\citeauthoryear{Redmon and Farhadi}{Redmon and
  Farhadi}{2017}]%
        {Redmon17}
\bibfield{author}{\bibinfo{person}{J. Redmon} {and} \bibinfo{person}{A.
  Farhadi}.} \bibinfo{year}{2017}\natexlab{}.
\newblock \showarticletitle{YOLO9000: Better, Faster, Stronger}.
\newblock \bibinfo{journal}{\emph{CVPR}} (\bibinfo{year}{2017}).
\newblock


\bibitem[\protect\citeauthoryear{Ren, He, Girshick, and Sun}{Ren
  et~al\mbox{.}}{2015}]%
        {Ren15}
\bibfield{author}{\bibinfo{person}{S. Ren}, \bibinfo{person}{K. He},
  \bibinfo{person}{R. Girshick}, {and} \bibinfo{person}{J. Sun}.}
  \bibinfo{year}{2015}\natexlab{}.
\newblock \showarticletitle{FASTER R-CNN: Towards Real-Time Object Detection
  with Region Proposal Networks}.
\newblock \bibinfo{journal}{\emph{NIPS}} (\bibinfo{year}{2015}).
\newblock


\bibitem[\protect\citeauthoryear{Salimans, Goodfellow, Zaremba, Cheung,
  Redford, and Chen}{Salimans et~al\mbox{.}}{2016}]%
        {Salimans16}
\bibfield{author}{\bibinfo{person}{T. Salimans}, \bibinfo{person}{I.
  Goodfellow}, \bibinfo{person}{W. Zaremba}, \bibinfo{person}{V. Cheung},
  \bibinfo{person}{A. Redford}, {and} \bibinfo{person}{X. Chen}.}
  \bibinfo{year}{2016}\natexlab{}.
\newblock \showarticletitle{Improved techniques for training GANs}.
\newblock \bibinfo{journal}{\emph{NIPS}} (\bibinfo{year}{2016}).
\newblock


\bibitem[\protect\citeauthoryear{Shakhnarovich}{Shakhnarovich}{2005}]%
        {Shakhnarovich05}
\bibfield{author}{\bibinfo{person}{G. Shakhnarovich}.}
  \bibinfo{year}{2005}\natexlab{}.
\newblock \showarticletitle{Learning Task-Specific Similarity}.
\newblock \bibinfo{journal}{\emph{MIT PhD Dissertation}}
  (\bibinfo{year}{2005}).
\newblock


\bibitem[\protect\citeauthoryear{Shakhnarovich, Viola, and
  Darrell}{Shakhnarovich et~al\mbox{.}}{2003}]%
        {Shakhnarovich03}
\bibfield{author}{\bibinfo{person}{G. Shakhnarovich}, \bibinfo{person}{P.
  Viola}, {and} \bibinfo{person}{T. Darrell}.} \bibinfo{year}{2003}\natexlab{}.
\newblock \showarticletitle{Fast Pose Estimation with Parameter Sensitive
  Hashing}.
\newblock \bibinfo{journal}{\emph{ICCV}} (\bibinfo{year}{2003}).
\newblock


\bibitem[\protect\citeauthoryear{Simonyan and Zisserman}{Simonyan and
  Zisserman}{2015}]%
        {Simonyan15}
\bibfield{author}{\bibinfo{person}{K. Simonyan} {and} \bibinfo{person}{A.
  Zisserman}.} \bibinfo{year}{2015}\natexlab{}.
\newblock \showarticletitle{Very Deep Convolutional Networks For Large-Scale
  Image Recognition}.
\newblock \bibinfo{journal}{\emph{ICLR}} (\bibinfo{year}{2015}).
\newblock


\bibitem[\protect\citeauthoryear{Yang, Luo, and Urtasun}{Yang
  et~al\mbox{.}}{2018}]%
        {Yang18}
\bibfield{author}{\bibinfo{person}{B. Yang}, \bibinfo{person}{W. Luo}, {and}
  \bibinfo{person}{R. Urtasun}.} \bibinfo{year}{2018}\natexlab{}.
\newblock \showarticletitle{{PIXOR}: Real-time {3D} Object Detection from Point
  Clouds}.
\newblock \bibinfo{journal}{\emph{CVPR}} (\bibinfo{year}{2018}).
\newblock


\bibitem[\protect\citeauthoryear{Yu and Koltun}{Yu and Koltun}{2016}]%
        {Yu16}
\bibfield{author}{\bibinfo{person}{F. Yu} {and} \bibinfo{person}{V. Koltun}.}
  \bibinfo{year}{2016}\natexlab{}.
\newblock \showarticletitle{Multi-Scale Context Aggregation by Dilated
  Convolutions}.
\newblock \bibinfo{journal}{\emph{ICLR}} (\bibinfo{year}{2016}).
\newblock


\bibitem[\protect\citeauthoryear{Zhang, Ren, and Urtasun}{Zhang
  et~al\mbox{.}}{2019}]%
        {Zhang19}
\bibfield{author}{\bibinfo{person}{C. Zhang}, \bibinfo{person}{M. Ren}, {and}
  \bibinfo{person}{R. Urtasun}.} \bibinfo{year}{2019}\natexlab{}.
\newblock \showarticletitle{Graph HyperNetworks for Neural Architecture
  Search}.
\newblock \bibinfo{journal}{\emph{ICLR}} (\bibinfo{year}{2019}).
\newblock


\bibitem[\protect\citeauthoryear{Zhou and Tuzel}{Zhou and Tuzel}{2018}]%
        {Zhou18}
\bibfield{author}{\bibinfo{person}{Y. Zhou} {and} \bibinfo{person}{O. Tuzel}.}
  \bibinfo{year}{2018}\natexlab{}.
\newblock \showarticletitle{{VoxelNet}: End-to-End Learning for Point Cloud
  Based {3D} Object Detection}.
\newblock \bibinfo{journal}{\emph{CVPR}} (\bibinfo{year}{2018}).
\newblock


\bibitem[\protect\citeauthoryear{Zoph and Le}{Zoph and Le}{2017}]%
        {Zoph17}
\bibfield{author}{\bibinfo{person}{B. Zoph} {and} \bibinfo{person}{Q.~V. Le}.}
  \bibinfo{year}{2017}\natexlab{}.
\newblock \showarticletitle{Neural architecture search with reinforcement
  learning}.
\newblock \bibinfo{journal}{\emph{ICLR}} (\bibinfo{year}{2017}).
\newblock


\bibitem[\protect\citeauthoryear{Zoph, Vasudevan, Shlens, and Le}{Zoph
  et~al\mbox{.}}{2018}]%
        {Zoph18}
\bibfield{author}{\bibinfo{person}{B. Zoph}, \bibinfo{person}{V. Vasudevan},
  \bibinfo{person}{J. Shlens}, {and} \bibinfo{person}{Q.~V. Le}.}
  \bibinfo{year}{2018}\natexlab{}.
\newblock \showarticletitle{Learning transferable architectures for scalable
  image recognition}.
\newblock \bibinfo{journal}{\emph{CVPR}} (\bibinfo{year}{2018}).
\newblock


\end{thebibliography}
